\documentclass[12pt]{article}
\usepackage{amsthm,amsmath,amsfonts,amssymb}
\usepackage{graphicx,psfrag,epsf,algorithm}
\usepackage{enumerate}
\usepackage{multirow} 
\usepackage[authoryear]{natbib}
\usepackage{url} % not crucial - just used below for the URL 
\usepackage{booktabs}
\usepackage{subcaption}
\usepackage{hyperref}
\usepackage[left]{lineno}
\usepackage{color}
\newcommand{\blind}{1}

\begin{document}
%{\tiny }\linenumbers % add line numbers
%\bibliographystyle{natbib}

\def\spacingset#1{\renewcommand{\baselinestretch}%
{#1}\small\normalsize} \spacingset{1}

%%%%%%%%%%%%%%%%%%%%%%%%%%%%%%%%%%%%%%%%%%%%%%%%%%%%%%%%%%%%%%%%%%%%%%%%%%%%%

\if0\blind
{
  \bigskip
  \bigskip
  \bigskip
  \title{\LARGE A Comparison of DeepSeek and Other LLMs}
%  \author{
%    Tianchen Gao, Jiashun Jin, Zheng Tracy Ke, and Gabriel Moryoussef \footnote{For contact, use jiashun@stat.cmu.edu or zke@fas.harvard.edu} 
% }
\date{}
  \maketitle
  \medskip
} \fi

\if1\blind
{
  \title{A comparison of DeepSeek and other LLMs}
  \author{Tianchen Gao \hspace{.2cm}\\ 
    Beijing International Center for Mathematical Research, Peking University\\  
    Jiashun Jin   \hspace{.2cm} \\
    Department of Statistics $\&$ Data Science, Carnegie Mellon University  \\ 
    Zheng Tracy Ke \hspace{.2cm} \\ 
    Department of Statistics, Harvard University  \\ 
    Gabriel Moryoussef \hspace{.2cm}  \\ 
    Department of Statistics $\&$ Data Science, Carnegie Mellon University}
  \maketitle
} \fi

\bigskip
\begin{abstract}
Recently,  DeepSeek has been the focus of attention in and beyond  the AI community. 
An interesting problem is how DeepSeek compares to other large language models (LLMs).   There are many tasks an LLM can do, and in this paper, 
we use the {\it task  of predicting an outcome 
using a short text} for comparison.  
We consider two settings, an authorship classification setting and a citation 
classification setting. In the first one, the goal is 
to determine whether a short text   is written by human or AI. 
In the second one, the goal is to classify a  citation  to one of  four types  
using the textual content.  
For each experiment, we compare DeepSeek with $4$ popular LLMs: Claude, Gemini, GPT, and Llama. 

\vspace{0.35 em} 

We find that,  in terms of classification accuracy,  DeepSeek outperforms Gemini, GPT, and Llama in most cases,  but underperforms Claude.  We also find that  DeepSeek is comparably slower than others but with a low cost to use, while 
Claude is much more expensive than all the others. 
Finally, we find that  in terms of similarity, the output  of DeepSeek is most similar to those of Gemini and Claude (and among all  $5$ LLMs, Claude and Gemini have the most similar outputs).  

\vspace{0.35 em}

In this paper,   we also present a fully-labeled dataset   collected by ourselves,   
and propose a recipe where we can use the LLMs and a  recent data set,  MADStat,  
to generate new data sets.   
The datasets in our paper can  be used as benchmarks for future study on LLMs. 
\end{abstract}

\noindent%
\vfill

\newpage
\spacingset{1.45} % DON'T change the spacing!

\section{Introduction} 
In early 2025, DeepSeek (DS),  a recent large language model (LLM) \citep{DeepSeek,deepseekai2025deepseekr1incentivizingreasoningcapability},   has shaken up the AI industry.  Since its latest version was released on January 20, 2025,  DS  has made the headlines of news and social media,  shot to the top of Apple Store's downloads, stunning investors and sinking some tech stocks including Nvidia.  

What makes DS so special is that in some benchmark tasks it achieved  the same or even better results as the big players in the AI industry (e.g., OpenAI's ChatGPT),  but with only a fraction of the training cost.  
For example, 
\begin{itemize} 
\item In \cite{DS-math}, the author showed that over $30$ challenging mathematical problems derived from the MATH dataset 
\citep{MATHdata}, DeepSeek-R1 achieves superior accuracy on these complex problems, compared with 
ChatGPT and Gemini, among others.  
\item  In a LinkedIn post on January 28, 2025, 
Javier Aguirre (a researcher specialized in medicine and AI,  South Korea) wrote:  ``I am quite impressed with Deepseek.  ....  
Today I had a really tricky and complex (coding) problem. Even chatGPT-o1 was not able to reason enough to solve it. I gave a try to Deepseek and it solved it at once and straight to the point".  This was echoed by several other researchers in AI. 
\end{itemize} 
See more comparison in \cite{DeepSeek, deepseekai2025deepseekr1incentivizingreasoningcapability,DeepSeek1, DeepSeek2}. 
 Of course, a sophisticated LLM has many aspects (e.g., Infrastructure,  Architecture,   Performances, Costs) and can achieve many tasks. The tasks discussed  above are only a small part of what an LLM can deliver.  It is desirable to have a more comprehensive and in-depth comparison. Seemingly, such a comparison may take a lot of time and efforts,  
but some interesting discussions  have  already appeared on the internet and social media (e.g.,  \cite{DS-comparison}).    

We are especially interested in how accurate an LLM is  in prediction. Despite a long list of statistical and machine learning algorithms for prediction \citep{ESL}, using LLM for prediction still has advantages: A classical approach may need a reasonably large training sample,  but an LLM can work with only a prompt. 
Along this line, a  problem of major interest is  how DS compares to  other LLMs in terms of prediction accuracy.     
In this paper, we consider two classification settings as follows. 
\begin{itemize} 
\item {\it Authorship classification (AC)}.   Determine whether a document is human-generated (hum),   or  AI-generated (AI), or human-generated but with AI-editing (humAI). 
\item {\it Citation classification (CC)}.  Given an academic citation and %the small piece of text surrounding it, 
the surrounding context, 
determine which type the citation is (see below for the $4$ pre-defined types). 
\end{itemize}  
There are several reasons why we select these two problems for our study.  First, these problems are of broad interests.  Detection of AI-generated content is an emerging hot topic \citep{liu2019roberta,ippolito2020automatic,solaiman2019release,mitchell2023detectgpt}. 
It enables the AI users to be aware of potential misinformation and incorrect advice in AI-generated content, 
especially for sensitive areas such as healthcare \citep{brown2020language}. 
Citation analysis is crucial for measuring researchers' scientific impact \citep{hirsch2005index}. 
For more accurate citation analysis, it is desirable to classify citations into refined categories \citep{teufel2006automatic,li2013towards,dong2011ensemble,pride2020authoritative}. 
Second,  these tasks do not require any specific knowledge, making it a fair comparison for LLMs trained for general purposes.  Last but not the least,  the two problems have the desirable difficulty level 
for comparing different LLMs: they are neither too easy nor too hard, with the error rates of different LLMs spanning a reasonably wide range.

%{\color{red}
%We selected two tasks that reflect important and complementary aspects of LLM performance.  
%Authorship classification addresses a timely real-world issue: how to detect AI-written or AI-edited content, which is central to concerns around misinformation, authorship verification, and transparency.  
%Citation classification, in contrast, requires domain-specific understanding of academic writing and the ability to interpret nuanced context.  
%Together, these tasks test both surface-level language generation (authorship) and deeper semantic understanding (citation use), giving a balanced picture of how different LLMs perform across a range of reasoning demands.
%}

For each of the two settings,  we compare the classification results of DeepSeek-R1 (DS)  with those of  $4$ representative LLMs: OpenAI's  GPT-4o-mini (GPT),  Google's Gemini-1.5-flash (Gemini), 
Meta's Llama-3.1-8b (Llama)  and Anthropic's Claude-3.5-sonnet (Claude).  We now discuss each of the two settings with more details.

\subsection{Authorship classification} 
\label{subsec:intro-AC}  
In the past two years,  AI-generated text content started to spread quickly, influencing the internet, workplace, and daily life.  This raises a problem: 
how to differentiate AI-authored content from human-authored content \citep{brown2020language,danilevsky2020survey}.

This problem is important for at least two reasons. First, AI-generated content can contain harmful misinformation in critical domains such as healthcare, news, and finance \citep{brown2020language},  potentially threatening the integrity of online resources. Second, identifying key differences between human- and AI-generated content can provide valuable insights for improving AI language models \citep{danilevsky2020survey}.

We approach the problem by considering  two classification settings, AC1 and AC2. 
\begin{itemize} 
\item {\it (AC1)}.  In the first setting, we focus on distinguishing human-generated text  and AI-generated text (i.e., hum vs. AI).  
\item {\it (AC2)}. In the second setting, we consider the more subtle setting of 
distinguishing text generated by human and text that are generated by human  but with AI editing (i.e., hum vs. humAI). 
\end{itemize} 
For experiments, we  propose to  use the recent MADStat data set  \citep{JBES, TextReview}.  MADStat is a large-scale data set on statistical publications, consisting of  the bibtex and citation information of 83,331 papers published in 36 journal in statistics and related field, spanning 1975-2015.  The data set is available for free download (see Section \ref{sec:main} for the download links). 

We propose a general recipe where we use the LLMs and MADStat to generate new data sets for our study. 
We start by selecting a few authors and collecting all papers authored by them in the MADStat. For each paper, the MADStat contains a title and an abstract. 
\begin{itemize} 
\item {\it (hum)}.  We use all the   abstracts as the  human-generated text.  
\item {\it (AI)}.   For each paper, we feed in the title to GPT-4o-mini and ask it to generate 
an abstract. We treat these abstracts as the AI-generated text. 
\item {\it (humAI)}.  For each paper, we also ask  GPT-4o-mini to edit the abstract. 
We treat  these abstracts as the humAI text.  
\end{itemize} 
%Seemingly, 
Using this recipe, we can generate many different data sets.   
These data sets provide a useful platform for us to compare 
different classification approaches, especially the $5$ LLMs. 

{\bf Remark 1} {\it (The MadStatAI data set)}.  In Section \ref{subsec:CC},  we fix $15$ authors in the MADStat data set  (see Table \ref{tb:15-author} for the list) and generate a data set containing 582 abstract triplets  (each triplet contains three abstracts: hum, AI, and humAI)  following the recipe above.  For simplicity, we call this  data set the {\it MadStatAI}.

Once the data set is ready, we apply each of the $5$ LLMs above for classification, 
with the same prompt. See Section \ref{subsec:AC} for details. 
Note that aside from LLMs,  we may apply other algorithms to this problem  \citep{solaiman2019release,zellers2019defending,gehrmann2019gltr,ippolito2020automatic,fagni2021tweepfake,adelani2020generating,alonAI,mitchell2023detectgpt}. However, since our focus in this paper is to compare DeepSeek with other LLMs, 
we only consider the $5$ LLM classifiers mentioned above.

\subsection{Citation classification} 
\label{subsec:intro-CC}  
When a paper is being cited, the citation could be significant or insignificant. 
Therefore, to evaluate this paper's impact, we are   interested in knowing not only  how many times it has been cited, 
but also how many of those citations are significant.  
While counting the total number of citations is relatively straightforward using tools like Google Scholar or Web of Science, identifying and quantifying `significant' citations remains a challenging task.

To address the issue, note that surrounding a citation instance, there is usually a short piece of text. The text contains important information for the citation, and we can use it to predict the type of this  citation.  
This gives rise to the problem of {\it Citation Classification}, where the goal is to use the short text surrounding a 
citation to predict the citation type.

We face two key challenges here. First, the types of different academic citations remain unclear. Second, there is no readily available dataset for analysis.

To address these challenges,  first,  after reviewing many literature works and empirical results,  we propose to divide all academic citations into four different types:  
\[
\begin{array}{l}
\mbox{``Fundamental ideas (FI)", \; ``Technical basis (TB)",}\\
\mbox{ ``Background (BG)", \;  and ``Comparison (CP)".}
\end{array}
\]
For convenience,  we encode the four types as ``1", ``2", ``3", ``4".  The first two types are considered as significant, and the other two  are considered as  
comparably less significant.  See Section \ref{subsec:CC} for details.

Second, with substantial efforts, we have {\it collected from scratch}  a new data set  by ourselves, which we call the {\it CitaStat}. 
We downloaded all papers in $4$ representative journals in statistics between 1996 and 2020 in PDF format. 
These papers contain about $360,000$ citation instances. For our study, we selected $n=3000$ citation instances. 
For each citation:  
\begin{itemize} 
\item We write a code  to select the small piece of text surrounding the citation in the PDF file and convert it to usable text files. 
\item We manually label each citation to each of the $4$ citation types above. 
\end{itemize} 
See Section \ref{subsec:CC} for details. 
As a result, CitaStat is a fully labeled data set with $n = 3000$ samples, 
where each $y$-variable takes values in $\{1, 2, 3, 4\}$ (see above), and each $x$-variable  
is a short text which we call the {\it textual content} of the corresponding citation. 

We can now use the data set to compare the $5$ LLMs above for their 
performances in citation classification.  We consider two experiments. 
\begin{itemize} 
\item {\it (CC1)}. A $4$-class classification experiment where we use the CitaStat without any modification.  
\item {\it (CC2)}. A $2$-class classification experiment where we merge class ``1" and ``2" (`FI' and `TB') to a new class 
of ``S" (significant), and we merge class ``3" and ``4" (`BG' and `CP') to a new class of `I' (incidental). 
\end{itemize}

\subsection{Results and contributions} \label{subsec:intro-results} 
We have applied all $5$ LLMs to the four experiments (AC1, AC2, CC1, CC2), and we have the following observations:  
\begin{itemize} 
\item In terms of classification errors,   Claude consistently outperforms all other LLM approaches.   DeepSeek-R1 underperforms Claude but outperforms Gemini, GPT, and Llama in most of the cases.  
GPT performs unsatisfactorily for AC1 and AC2,  with an error rate similar to that of random guessing, but it performs much better than random guessing for CC1 and CC2.  Llama performs unsatisfactorily: its error rates are either comparable to those of random 
guessing or even larger.   
\item In terms of computing time,  Gemini and GPT are much faster than the other three approaches, and DeepSeek-R1 
is the slowest (an older version of DeepSeek, DeepSeek-V3,   is faster but does not perform as well as DeepSeek-R1). 
\item In terms of cost, Claude is  much more expensive for a customer than other approaches.  For example, for CC1 and CC2 altogether, Claude costs $\$12.30$, Llama costs $\$1.2$, and the other three methods (DeepSeek, Gemini and GPT) cost no more than $\$0.3$. 
\item In terms of output similarity,   DeepSeek is most similar to Gemini and Claude (GPT and Llama are highly similar in AC1 and AC2, but both perform relatively unsatisfactorily). 
\end{itemize} 
Table \ref{table:ranking} presents  the ranks of different approaches in terms of error rates (the method with the lowest error rate is assigned a rank of $1$).   The average ranks suggest that DeepSeek outperforms Gemini, GPT, and Llama, but 
underperforms Claude (note that for CC1 and CC2, we have used two versions of DeepSeek, R1 and V3; 
the results in Table \ref{table:ranking} are based on R1. If we use V3, then 
DeepSeek ties with Gemini in average rank; it still  outperforms GPT and Llama). See Section 
\ref{sec:main} for details.

\spacingset{1.2} 
%%%%%%%%%
%%%%%%%%%
%%%%%%%%%
\begin{table}[htb!]
\centering
\scalebox{1}{
\begin{tabular}{lccccc}
\toprule
  &Claude  & DeepSeek & Gemini & GPT & Llama \\   
  \midrule
Experiment AC1 (hum vs. AI)   &1 &2  &3  &5 &4    \\ 
Experiment AC2 (hum vs. humAI) & 2 & 1& 3 & 5 & 4 \\ 
Experiment CC1 (4-class)  & 1 &4 & 2& 3 & 5 \\ 
Experiment CC2 (2-calss)  & 1 &2 & 3& 4 & 5  \\  
\midrule
Average Ranking & 1.25  &2.25  &2.75  & 4.25  & 4.50  \\ 
\bottomrule
\end{tabular}}
\caption{The rankings of all $5$ LLM approaches in terms of error rates.}
\label{table:ranking}
\end{table}
\spacingset{1.45}

Overall, we find that Claude and DeepSeek have the lowest error rates, but 
Claude is relatively expensive and DeepSeek is relatively slow.

Additionally, we have investigated the temporal stability of LLMs, the comparison with traditional classification methods, and  the use of different LLMs for generating AI-content.  
\begin{itemize}
\item In terms of short-term and long-term variations across multiple runs, Claude, DeepSeek, and Gemini are stable (with more than 90\% of agreement on the predicted labels between two separate runs, in the CC1 and CC2 experiments), GPT is moderately stable (with 80-90\% agreement), and Llama is  unstable (with around 50\% agreement). 
\item A traditional classification method, the Higher Criticism classifier \citep{DJ08} significantly outperforms all the LLM approaches. Inspired by this observation, we propose a hybrid classifier that combines the strengths of HC and an LLM. In the AC1 experiment, the hybrid of HC and Claude achieves the lowest error rate among all classifiers considered in this paper.  
\item Note that we use one LLM to generate the AI content and another LLM to perform classification, and 
an interesting question is whether it is critical to use the same LLM in the two components.  
For classification without feature selection by HC, this is not critical. 
However, for classification with feature selection by HC, we do observe 
a much better result  when we use the same LLM in the two components. 
\end{itemize}

We have made the following contributions.  First, as DeepSeek 
has been the focus of attention in and beyond the  AI community,   there is a timely need to understand how 
it  compares  to other popular LLMs.  Using two interesting classification problems, 
we demonstrate that DeepSeek is competitive in the task of predicting an outcome using a short piece of text.  
Second, we  propose citation classification as an interesting new problem, the understanding of which 
will help evaluate the impact of academic research. 
Last but not the least,  we provide CitaStat as a new data set which can be used for evaluating academic research. We also propose a general recipe for generating new data sets (with the MadStatAI as an example) for studying AI-generated text. 
These data sets can be used as benchmarks to compare 
different algorithms, and  to learn the differences between 
human-generated text and AI-generated text. 

%\vspace{-.5cm}
 
%%%%%%%%%%%%%%%
%%%%%%%%%%%%%%%    
\section{Main results} 
\label{sec:main} 
In Sections~\ref{subsec:AC}-\ref{subsec:CC}, we describe our experiments on authorship classification and citation classification, and report the results. 
In Section~\ref{subsec:stability}, we investigate the temporal stability of LLMs. 
In Section~\ref{subsec:HC}, we compare the performances of LLMs with those of statistical classification approaches, and propose a way for combining the strengths of both. 

%\vspace{-.5cm}

\subsection{Authorship classification} 
\label{subsec:AC} 

The MADStat  \citep{JBES,TextReview} contains over 83K  abstracts, but it is time-consuming to process all  of them.\footnote{MADStat stands for Multi-Attributed Dataset on Statisticians.  MADStat is available 
for free download at \url{http://tracyke.net/MADStat.html} or in the \href{https://doi.org/10.1080/07350015.2021.1978469}{supplementary material} of \cite{JBES}.}
 We selected a small subset as follows: First, we restricted to authors who had over 30 papers in MADStat. Second, we randomly drew 15 authors from this pool by sampling without replacement. Each time a new author was selected, we checked if he/she had co-authored papers with previously drawn authors; if so, we deleted this author and drew a new one, until the total number of authors reached 15. Finally, we collected all 15 authors' abstracts in MADStat. This gave rise to a data set with 582 abstracts in total (see Table~\ref{tb:15-author}).

\spacingset{1.2}
\begin{table}[htb!]
\centering
\scalebox{.9}{
\begin{tabular}{lc|lc|lc}
\toprule
Name & \#abstracts & Name & \#abstracts & Name & \#abstracts\\
\midrule
Andrew Gelman & 40 & Anthony Pettitt & 60 & Damaraju Raghavarao &31\\
David Nott & 35 & Frank Proschan &39 & Ishwar Basawa &53\\
Ngai Hang Chan &32 & Nicholas I. Fisher & 32  & Peter X. K. Song & 32\\
Philippe Vieu & 31 &Piet Groeneboom & 30 & Richard Simon & 45\\
Sanat K. Sarkar & 44  & Stephane Girard & 33 & Yuehua Wu & 45 \\
\bottomrule
\end{tabular}}
\caption{The 15 selected authors and their numbers of abstracts in MADStat.} \label{tb:15-author}
\end{table} 
\spacingset{1.45}

\begin{figure}[h!]
  \centering
    \includegraphics[width=.49\textwidth]{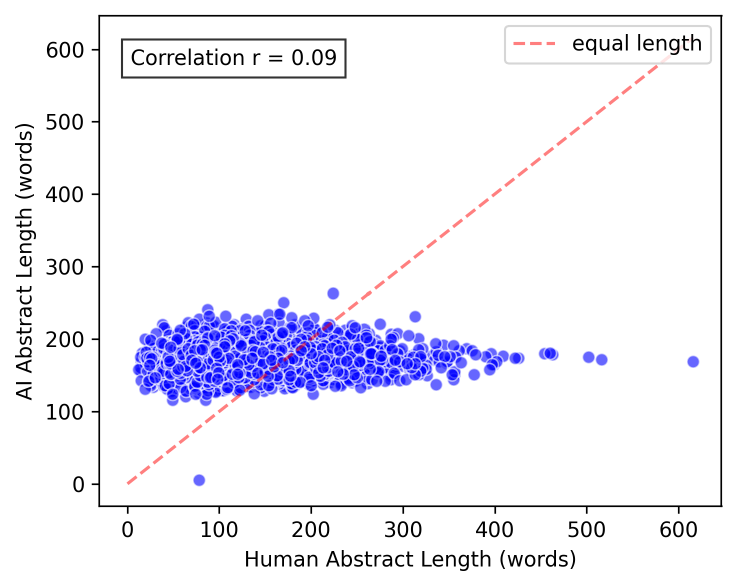}
    \includegraphics[width=.49\textwidth]{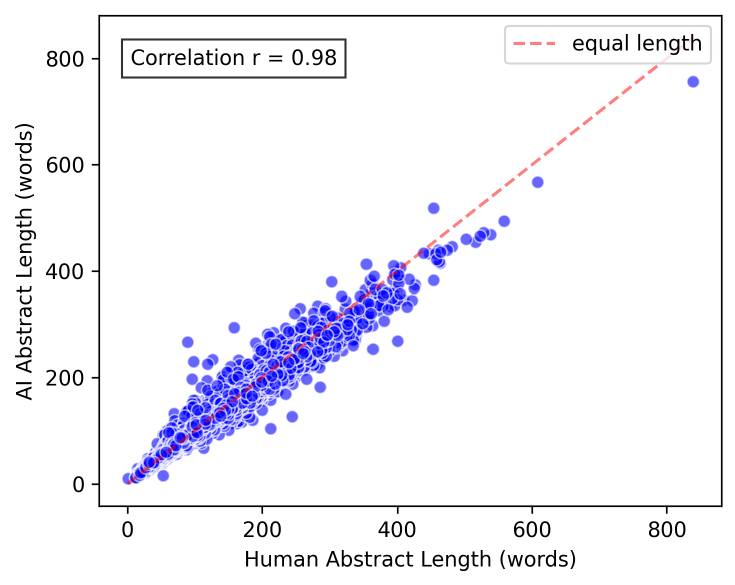}
  \caption{Comparison of the lengths of  human-generated and AI-generated abstracts. The x-axis is the length of an original abstract, and the y-axis is the length of its AI counterpart (left panel) or humAI counterpart (right panel).}
\label{fig:length}
\end{figure}

For each original human-written abstract, we used GPT-4o-mini to get two variants. 
\begin{itemize}
\item The AI version. We provided the paper title and requested for a new abstract. The prompt is \textit{``Write an abstract for a statistical paper with this title: [paper title]."} 

\item The humAI version. We provided the original abstract and requested for an edited version. The prompt is \textit{``Given the following abstract, make some revisions. Make sure not to change the length too much. [original abstract]."}
\end{itemize}
Both variants are authored by AI, but  they differ in appearance. The AI version is often significantly different from the original abstract, so the `human versus AI' classification problem is easier. 
For example, the left panel of Figure~\ref{fig:length} is a comparison of the length of the original abstract with that of its AI version. The length of human-written abstracts varies widely, while the length of AI-generated ones is mostly in the range of 100-200 words. The humAI version is much closer to the original abstract, typically only having local word replacements and mild sentence re-structuring. In particular, its length is highly correlated with the original length, which can be seen in the right panel of Figure~\ref{fig:length}.

\spacingset{1.1}
\spacingset{1.45}

As mentioned,  we consider two classification problems:
\begin{itemize} 
\item (AC1).   A $2$-class classification problem of `human versus AI', 
\item (AC2).   A $2$-class classification problem of  `human versus humAI'. 
\end{itemize}  
For each problem, there are $582\times2 = 1164$ testing samples, half from each class.  We input them into each of the 5 LLMs using the same prompt: \textit{``You are a classifier that determines whether text is human-written or AI-edited. Respond with exactly one word: either `human' for human-written text or `ChatGPT' for AI-written text. Be as accurate as possible."}

Note that comparing with classification approaches (e.g., SVM, Random Forest \citep{ESL}), 
an advantage of using an LLM for classification is that, we do not need to provide any training sample.  
All we need is to input the LLM with a prompt.

Table~\ref{table:accuracy_llms} summarizes the performances of 5 LLMs. 
For `human vs AI' (AC1), Claude-3.5-sonnet has the best error rate 0.218, and DeepSeek-R1 has the second best one 0.286. 
The remaining three methods almost always predict `human-written', which explains why their error rates are close to 0.5. 
For `human vs humAI' (AC2), since the problem is harder, the best achievable error rate is much higher than that of `human vs AI' (AC1).  DeepSeek-R1 has the lowest error rate 0.405, and Claude-3.5-sonnet has the second best one 0.435. The error rates of the other three methods are nearly 0.5. In conclusion, Claude-3.5-sonnet and DeepSeek-R1 are the winners in terms of error rate. If we also take the running time into account, Claude-3.5-sonnet has the best overall performance.  On the other hand, the cost of Claude-3.5-sonnet is the highest.

\spacingset{1.2}
\begin{table}[t!]
\centering
\scalebox{1}{
\begin{tabular}{l | c r c | c r c}
\toprule
\multirow{2}{*}{\textbf{Method}} & \multicolumn{3}{c|}{human vs. AI} & \multicolumn{3}{c}{human vs. humAI}\\
\cline{2-7}
& \textbf{Error} & \textbf{Runtime} & \textbf{Cost }& \textbf{Error} & \textbf{Runtime} & \textbf{Cost}\\
\hline
Claude-3.5-sonnet & 0.218  & 7 min & \$ 0.5 USD & 0.435 & 7 min &\$ 0.3 USD\\
DeepSeek-R1 & 0.286  & 235 min & \$ 0.05 USD &  0.405 & 183 min & \$ 0.04 USD\\
Gemini-1.5-flash & 0.468 & 6 min &\$ 0.1 USD & 0.500 & 6 min &\$ 0.09 USD \\
GPT-4o-mini & 0.511 & 7 min &\$ 0.1 USD & 0.502 & 8 min &\$ 0.12 USD\\
Llama-3.1-8b &  0.511 & 11 min & \$ 0.2 USD& 0.501 & 12 min &\$ 0.17 USD\\
\bottomrule
\end{tabular}}
\caption{The classification errors, runtime, and costs of 5 LLMs for authorship classification. (In `human vs AI' (AC1), if we report four digits after the decimal point, Llama-3.1-8b has a lower error than GPT-4o-mini. This is why they have different ranks for AC1 in Table~\ref{table:ranking}.)}
\label{table:accuracy_llms}
\end{table}
\spacingset{1.45}

{\bf Remark 2}: The relatively long runtime of DeepSeek-R1 is primarily due to its design. Compared to DeepSeek-V3 (the version of DeepSeek without reasoning) and 
other models, DeepSeek-R1 incorporates more complex reasoning chains, multi-step verification processes, and additional internal deliberation layers. While these design choices enhance the model’s accuracy, they also introduce significant computational overhead, making DeepSeek-R1 less practical for time-sensitive or high-throughput applications.

\begin{figure}[htb!]
    \centering
    \includegraphics[width=0.85\textwidth]{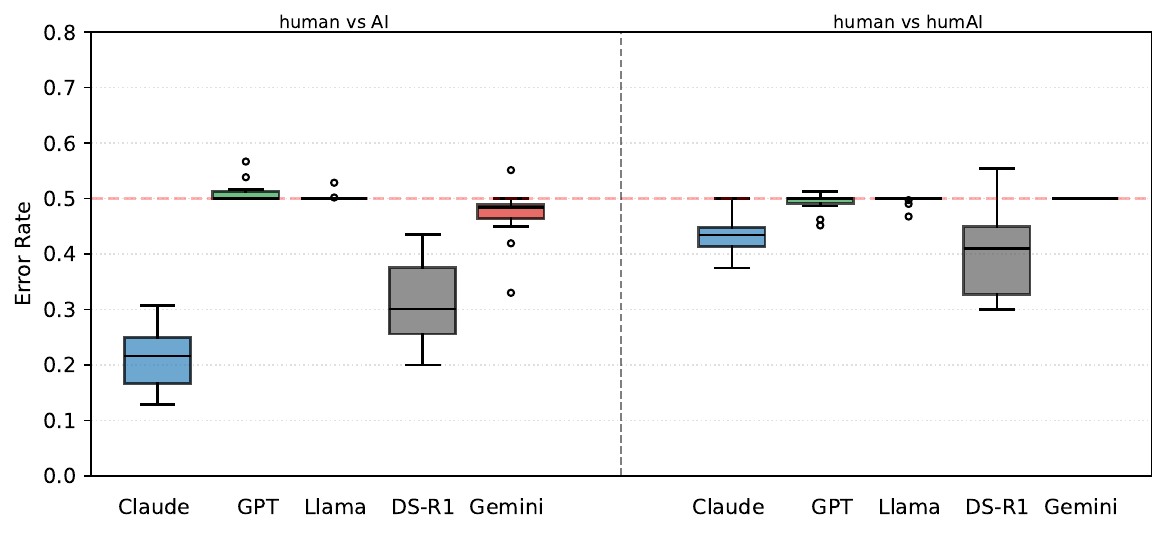}
    \caption{The boxplots of per-author classification errors.}
    \label{fig:boxplot}
\end{figure}

Since the 1164 testing abstracts come from 15 authors, we also report the classification error for each author (i.e., the testing documents only include this author's human-written abstracts and the AI-generated variants). 
Figure~\ref{fig:boxplot} shows the boxplots of per-author errors for each of 5 LLMs. 
Since authors have different writing styles, these plots give more information than Table~\ref{table:accuracy_llms}. For `human vs AI' (AC1), Claude-3.5-sonnet is still the clear winner. For `human vs humAI' (AC2),  DeepSeek-R1 still has the best performance. Furthermore, its advantage over Claude-3.5-sonnet is more visible in these boxplots: Although the overall error rates of two methods are only mildly different, DeepSeek-R1 does have a significantly better performance for some authors.

\spacingset{1.2}
\spacingset{1.45}

\begin{figure}[!ht]
	\centering
	\includegraphics[height=.439\textwidth, trim=0 0 100 0, clip=true]{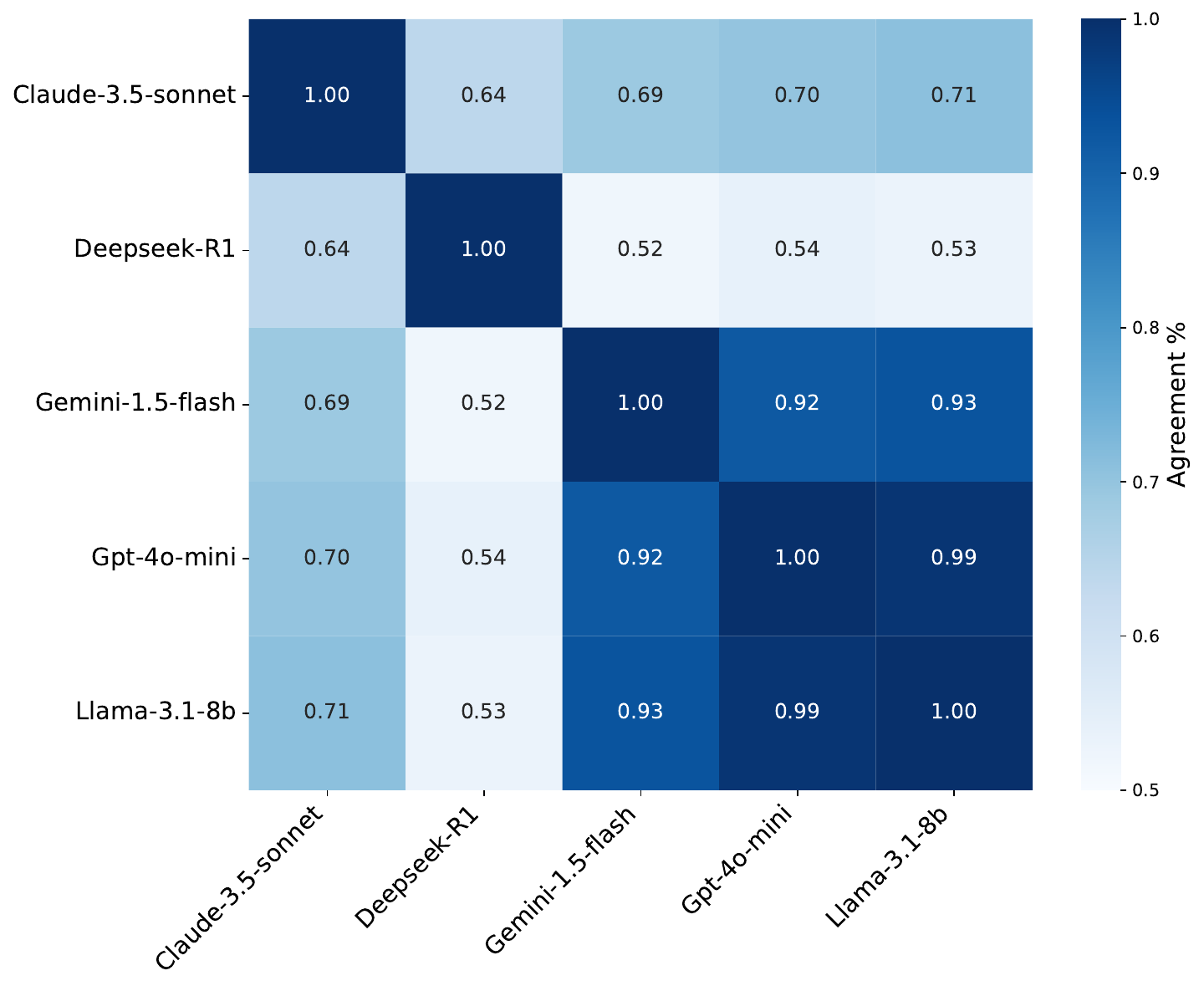} 
	\includegraphics[height=.439\textwidth]{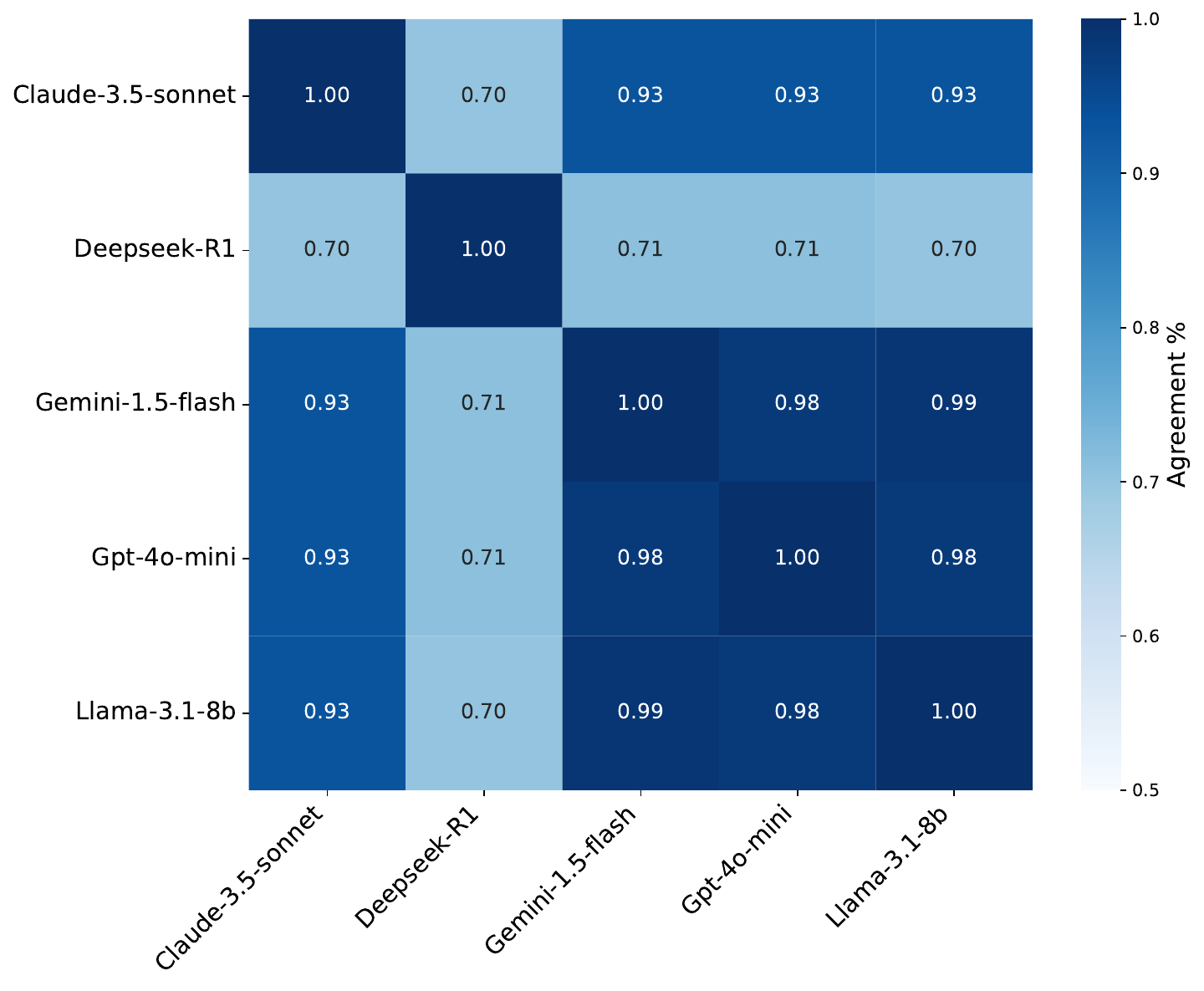}
	\caption{The prediction agreement among 5 LLMs in detecting AI from human texts. Left: `human versus AI' (AC1). Right: 
	`human versus humAI' (AC2). Take the cell on the first row and second column (left panel) for example: 
	for $64\%$ of the samples, the predicted outcomes by Claude-3.5-sonnet and DeepSeek-R1 are exactly the same.} 
	\label{fig:agreement-humanai}
\end{figure}

We also investigate the similarity of predictions made by different LLMs.  
For each pair of LLMs, we calculate the percent of agreement on predicted labels, in both the `human versus AI' (AC1)  setting and `human versus humAI' (AC2) settings. The results are in Figure~\ref{fig:agreement-humanai}. 
For both settings,  Gemini-1.5-flash, GPT-4o-mini, and Llama-3.1-8b have extremely high agreements with each other. 
This is because all three models predict `human-written' for the majority of samples. DeepSeek-R1 and Claude-3.5-sonnet are different from the other three, and they have 64\% and 70\% agreements with each other in two settings, respectively.

\spacingset{1.2}
%%%%%%%%%%%%%%%%%%%%%%%
\spacingset{1.45}

 {\bf Remark 3}. We briefly explain why an LLM succeeds (or fails) on the above task.  
Existing studies have shown that AI-generated text and human-written text  have different characteristics (e.g., \cite{reinhart2025llms}). For instance, an LLM may use certain words or sentence structures more frequently than human. Such differences can be detected by analyses of word frequencies or sentence-level likelihoods.
In our experiments,  especially in AC2,  we find that word frequencies contain important signals for distinguishing the AI-generated and human-generated texts  (see also Section~\ref{subsec:HC}).  In light of such an observation,  we conjecture that an LLM performs well on our tasks for its internal layers activated by the prompt capture some important information of the word frequencies.

 {\bf Remark 4} {\it (Generation of AI content)}:  
To obtain AI-generated content, a popular approach is to pick 
an LLM, feed it with a small number of words of a document (called the initial words),  and then ask it  to produce a complete document \citep{jawahar2020automatic}.  The approach we use  GPT-4o-mini to generate paper abstracts follows a similar spirit, where we treat the paper title as the ``initial words."  Alternatively, we may feed the whole paper to the LLM and ask it to generate the abstract.  
As the LLM gets to see more information, this approach may produce an abstract  closer to the human-written one.   
However, such an approach is practically infeasible: Most papers in MADStat are long, 
and it may exceed the token limit of the LLM even if we only include the introduction of the paper. 
For this reason, we stick to our current approach above.  While our focus is on 
abstracts generated by GPT-4o-mini, we also investigated  abstracts generated by other LLMs;  see Section~\ref{subsec:cross-LLM}.

\subsection{Citation classification} 
\label{subsec:CC} 
The MADStat only contains meta-information and abstracts, rather than full papers. We created a new dataset, CitaStat, by downloading full papers and extracting the text surrounding citations. 
Specifically, we restricted to those papers published during 1996-2020 in 4 representative journals in statistics: {\it Annals of Statistics}, {\it Biometrika}, {\it Journal of the American Statistical Association}, and {\it Journal of the Royal Statistical Society Series B}. 
We wrote code to acquire PDFs from journal websites, convert PDFs to plain text files, and extract the sentence (using SpaCy, a well-known python package for sentence parsing) containing each citation (we call it a `citation instance'). There are over 367K citation instances in total. We randomly selected $n=3000$ of them and manually labeled them using one of the following four categories: 
\begin{itemize}
\item  `Fundamental Idea (FI)' (previous work that directly inspired or provided important ideas for the current paper). Example:  
			{\it ``The proposed class of discrete transformation survival models is originally motivated from the continuous generalized odds-rate model by Dabrowska and Doskum (1988a) with time-invarying covariate and Zeng and Lin (2006) for..."}
\item `Technical Basis (TB)' (important tools, methods, data sets, and other resources).  Example: {\it  ``We solve this system numerically via the Euler method (Protter and Talay 1997;Jacod 2004) with a time-step of one day..."}
\item `Background (BG)' (background, motivations, related studies, and examples to support/illustrate points).  Example: {\it ``Estimation of current and future cancer mortality broken down by geographic area (state) and tumor has been discussed in a number of recent articles, including those by Tiwari et al. (2004) ..."} 
\item `Comparison (CP)'  (comparison of methods or theoretical results). Example: {\it ``Another way of determining the number of neuron pairs is to follow Medeiros and Veiga (2000b) and Medeiros et al. (2002) and use a sequence of..."}
\end{itemize}
These definitions were inspired by existing studies on citation types \citep{moravcsik1975some,teufel2006automatic,dong2011ensemble}. 
Occasionally, two categories may overlap. For example, a reference is cited as providing a fundamental idea, and a comparison with it is also stated in the same sentence. In this case, we label it as `Fundamental Idea (FI)', to highlight that this is more important than a general comparison.  In addition, TB and FI are considered important citations—either the citing paper directly uses the method from the cited work (TB), or the cited work inspires the core contribution of the citing paper (FI). In contrast, CP and BG do not require the cited work to play a critical role in the citing paper. For example, in the sentence ``Another way of determining the number of neuron pairs is to follow Medeiros and Veiga (2000b) and Medeiros et al. (2002)",  there is no indication that the methodology in the citing paper is inspired by the cited work and   the cited work is only mentioned as alternative approaches to compare. This aligns with our definition of CP, and thus we label it as CP rather than TB.

There were 20 citation instances for which manual labeling gave `not sure'. We removed them and obtained $n=2980$ labeled samples (see Table~\ref{tb:CitaStat}). 
\spacingset{1.2}
%%%%%%%%%%%%%%%%%%%%%%%%%%%%%%%
\begin{table}[tb!]
\centering
\scalebox{.95}{
\begin{tabular}{c|cccc|c}
\toprule
& Background & Comparison & Fundamental Idea & Technical Basis & Total\\
\midrule
Count & 1721 & 316 & 169 & 774 & 2980\\ 
Fraction & 57.8\% & 10.6\% & 5.7\% & 26\% & 100\% \\
\bottomrule
\end{tabular}}
\caption{The distribution of four classes in CitaStat.} \label{tb:CitaStat}
\end{table}
\spacingset{1.45}

With this CitaStat dataset, we consider two problems as mentioned:
\begin{itemize}
\item (CC1). The 4-class classification problem: Given the textual content of a citation (i.e., the text surrounding the  citation),  we aim to classify it to one of the four classes. 
\item (CC2).  The 2-class classification problem: We re-combine the four classes into two, where `Fundamental Idea'  and `Technical Basis'  are considered as `Significant (S)', and `Background' and `Comparison' are considered as `Incidental (I)'.\spacingset{1}\footnote{As mentioned, when a citation potentially belongs to two categories (e.g., `Fundamental Idea' and `Comparison'), we always manually label it to the more `significant' one (e.g., `Fundamental Idea'). This prevents mis-interpreting important comparisons as `incidental' citations in the manual labeling.}\spacingset{1.45}
Given the textual content of a citation, we aim to predict whether it is a `Significant (S)' citation. 
\end{itemize}

We used prompts to get classification decisions. Unlike the previous authorship classification problem, the class definitions in this problem are not common knowledge and need to be included in the prompt. In the 2-class problem, we use the prompt as in Figure~\ref{fig:prompt-CC}.  It provides definitions, examples, and how the four classes are re-combined into two, aiming to convey to the LLM as much information as possible. The prompt for the 4-class problem is similar, except that the description of grouping 4 classes into 2 is removed and the requirement for output format is modified (see details in the caption of Figure~\ref{fig:prompt-CC}).

\spacingset{1.1}
\begin{figure}[tb!]
\centering
\scalebox{.9}{
\fbox{
\begin{minipage}{1\textwidth}
{\it The content in the text comes from a paragraph in an academic paper A that includes citations. Please classify the citation [Reference Key] appearing in the following text into one of the categories: 
\begin{itemize} \itemsep 0pt
\item Background (citations that include descriptions of the research background, summaries of previous or recent studies and methods in a general way, and examples to support and illustrate points. For example, [Example 1], 
\item Comparison (citations that compare methods or results with those of this paper. For example, [Example 2], 
\item Fundamental idea (citations about the previous work that inspired or provided ideas for this paper. For Example, [Example 3], 
\item Technical basis (citations of important tools, methods, data, and other resources used in this paper. For example, [Example 4]. 
\end{itemize}
Furthermore, we consider Background or Comparison as Incidental, and Fundamental idea or Technical basis as Important.

\medskip

Text: [Citation text]

\medskip

Please reply only with one of the following: Important or Incidental.}
\end{minipage}}}
\caption{The prompt for 2-class citation classification, where [Reference Key] is the phrase in the text representing this reference, and [Example 1] is an example text from Background (other categories are similar). The prompt for 4-class classification is similar, except that the sentence {\it ``Furthermore, we consider ..."} is removed and the last sentence is changed to {\it ``Please reply only with one of the following: Background, Comparison, Fundamental idea, or Technical basis."}
} \label{fig:prompt-CC}
\end{figure}
\spacingset{1.45}

We examined the performances of all 5 LLMs. Since the runtime of DeepSeek-R1 is much longer than other methods, we also ran DeepSeek-V3, an earlier-released version of DeepSeek-R1, to evaluate its classification error rate. The results are reported in Table~\ref{tb:CC-results}.

For the 4-class problem, Claude-3.5-sonnet achieves the lowest error rate at 0.327, followed by Gemini-1.5-flash (0.347) and GPT-4o-mini (0.363). DeepSeek-R1 performs moderately, with an error of 0.380, which is better than DeepSeek-V3 (0.432) and Llama-3.1-8b (0.576) but inferior to the other three models. For the 2-class problem, Claude-3.5-sonnet still has the best error rate (0.261). DeepSeek-R1 is the second best (0.291), outperforming Gemini-1.5-flash (0.313), DeepSeek-V3 (0.332), and GPT-4o-mini (0.371).  In summary, Claude-3.5-sonnet achieves the best performance on both problems. DeepSeek-R1, although not the best, shows notable improvement over DeepSeek-V3.

Regarding the runtime, GPT-4o-mini and Gemini-1.5-flash are the fastest, taking only 15 and 25 minutes respectively. DeepSeek-V3 and Llama-3.1-8b require several hours. DeepSeek-R1 is the slowest, taking 60-65 hours (reasons of the slow runtime of DeepSeek-R1 have been discussed in Remark~2). Regarding the cost, DeepSeek-V3 is the cheapest, and Claude-3.5-sonnet is significantly more expensive than the other methods.

\spacingset{1.2}
\begin{table}[!ht]
	\centering
	\scalebox{1}{
		\begin{tabular}{lccll}
			\toprule
			\textbf{Model} & \textbf{Error (4-class)} & \textbf{Error (2-class)} & \textbf{Runtime} & \textbf{Cost} \\
			\midrule
			Claude-3.5-sonnet & 0.327 & 0.261 & 1-2 h & 12.30 USD \\
			DeepSeek-V3 & 0.432 & 0.332 & 3-4 h & 0.60 RMB \\
			 DeepSeek-R1 &  0.380  &  0.291 & 60-65 h  &   11.45 RMB  \\
			Gemini-1.5-flash & 0.347 & 0.313 & 25 min & 0.12 USD \\
			GPT-4o-mini & 0.363 & 0.371 & 15 min & 0.30 USD  \\
			Llama-3.1-8b & 0.576 & 0.457 & 4-5 h & 1.20 USD \\
			\bottomrule
	\end{tabular}}
	\caption{The error rate, runtime and cost of 6 LLMs for citation classification. } 	\label{tb:CC-results}
\end{table}
\spacingset{1.45}

Additionally, in the 4-class setting, we divided all samples into three groups according to the average prediction error by 6 LLMs (i.e., average of 6 binary values). The lowest 30\%, middle 40\%, and highest 30\% are called the Easy, Medium, and Difficult case, respectively. Table~\ref{tb:easydifficult} shows the per-group error rates of all 6 LLMs. In the Easy case, the error rates of all methods except Llama-3.1-8b are less than 0.01. 
In the Difficult case, all methods perform poorly, with GPT-4o-mini attaining the smallest error rate at 0.829 and DeepSeek-V3 attaining the biggest at 0.951.  In the Medium case, Claude-3.5-sonnet performs well with an error rate of 0.181, followed by Gemini-1.5-flash with a similar error rate 0.211. Llama-3.1-8b shows a notably higher error rate of 0.727. Compared to DeepSeek-V3, DeepSeek-R1 performs better in all three cases.

\spacingset{1.1}
\begin{table}[!ht]
	\centering{
		\scalebox{1}{
			\begin{tabular}{lccc}
				\toprule
				Model & Easy & Medium & Difficult \\
				\hline
				Claude-3.5-sonnet & .006 & .181 & .841 \\
				DeepSeek-V3 & .004 & .364 & .951 \\
				DeepSeek-R1 & .003 & .236 & .947 \\
				Gemini-1.5-flash & .006 & .211 & .869 \\
				GPT-4o-mini & .009 & .280 & .829 \\
				Llama-3.1-8b & .077 & .727 & .872 \\
				\bottomrule
		\end{tabular}}
		\caption{The per-group error rates for 6 LLMs in citation classification.} 	\label{tb:easydifficult}}
\end{table}
\spacingset{1.45}

\begin{figure}[htb]
	\centering
	\includegraphics[height=.436\textwidth, trim= 0 0 70 0, clip=true]{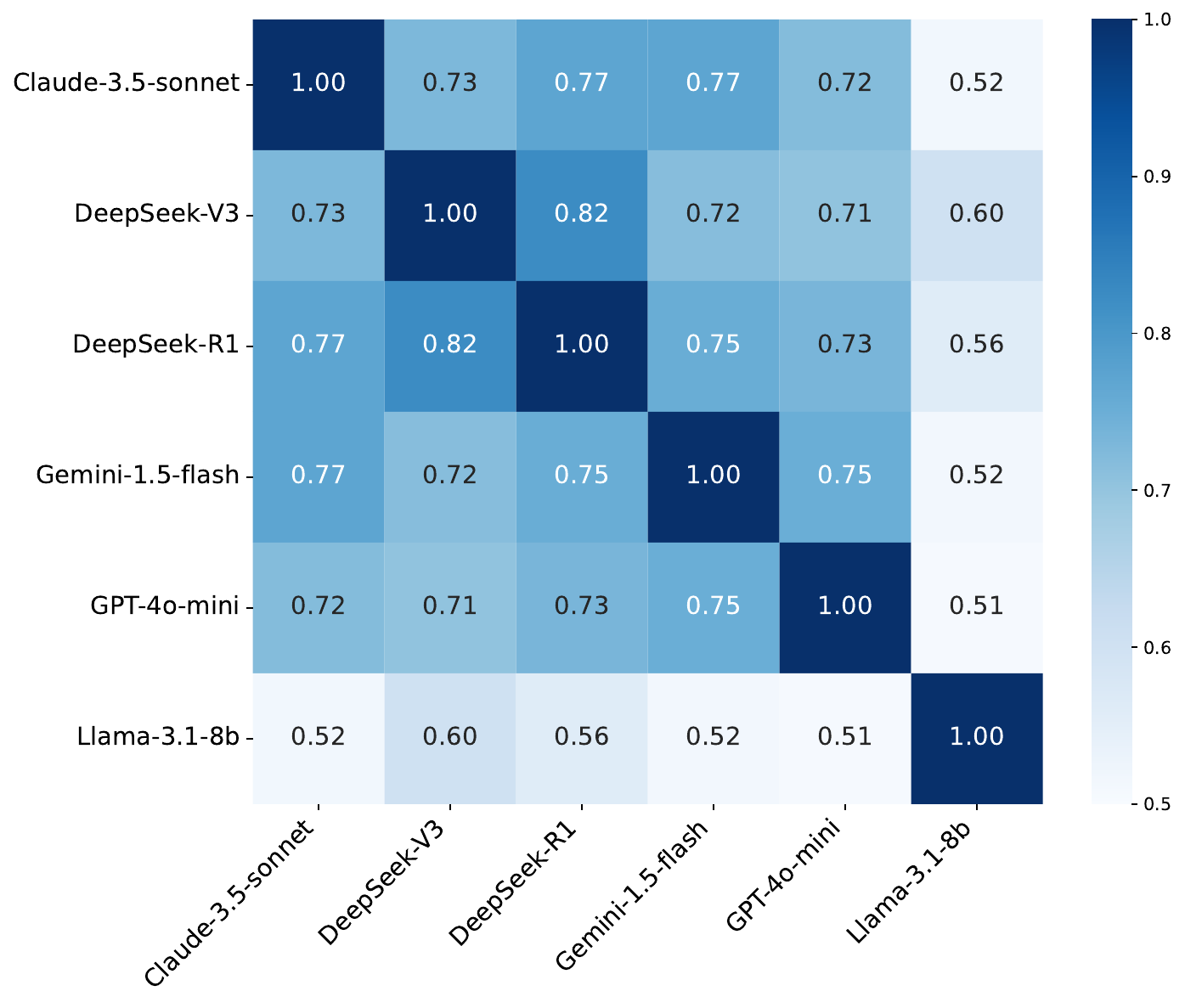} 
	\includegraphics[height=.436\textwidth]{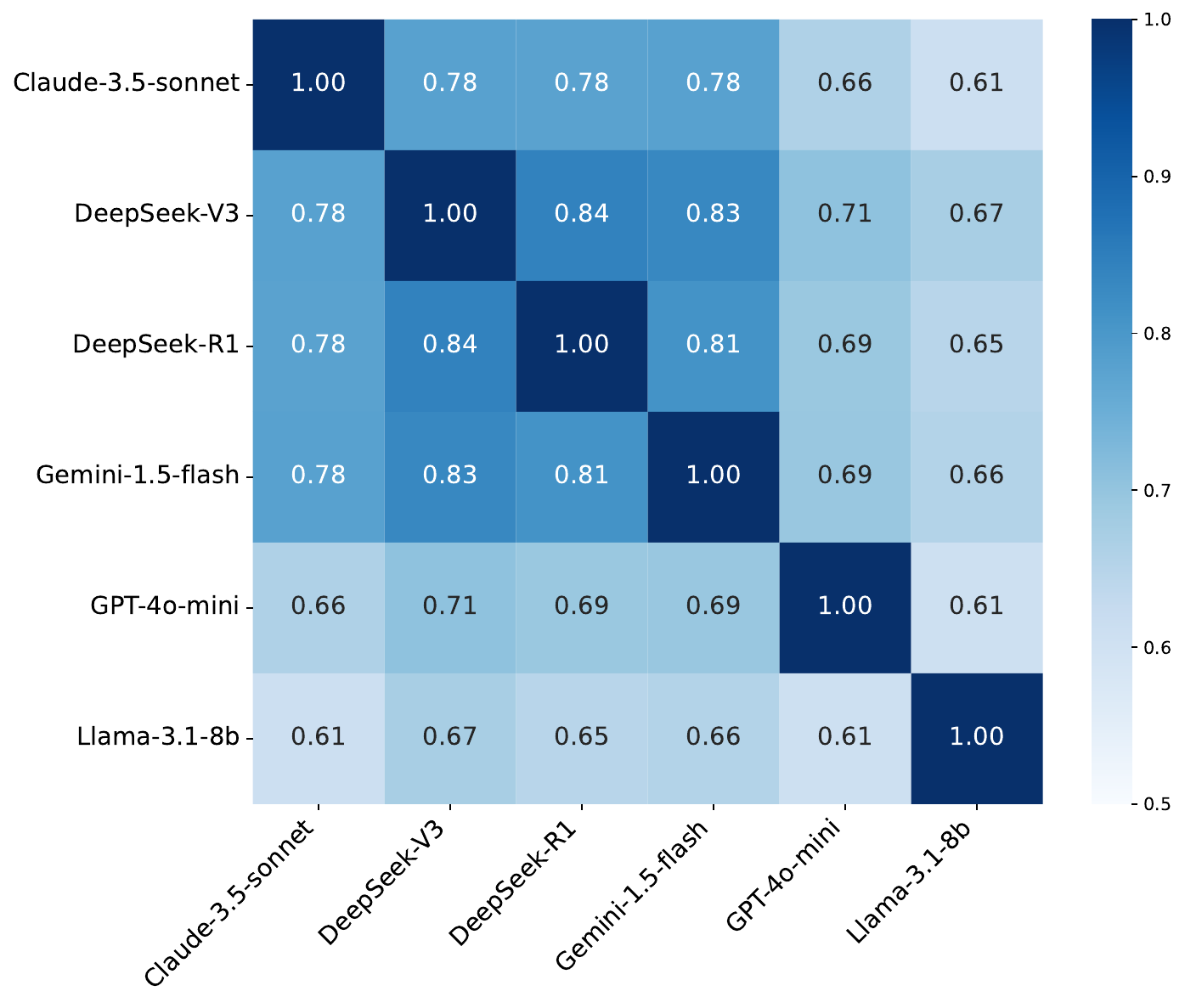}
	\caption{The prediction agreement among 6 LLMs in citation classification. Left: 4-class citation classification  (CC1). Right: 
	2-class citation classification (CC2). Take the cell on the first row and second column (left panel) for example: 
	for $73\%$ of the samples, the predicted outcomes by Claude-3.5-sonnet and DeepSeek-V3 are exactly the same.}
	\label{fig:agreement}
\end{figure}

Finally, we investigate the similarity of predictions made by different LLMs. 
For each pair of the 6 LLMs, we calculated the percent of agreement on predicted labels, for both the 4-class and 2-class settings. The results are given in Figure~\ref{fig:agreement}.

Noting that DeepSeek-V3 and DeepSeek-R1 are only different versions of the same LLM, we first exclude DeepSeek-V3 and check the agreement among the remaining 5 LLMs. 
In the 4-class setting (CC1), except for Llama-3.1-8b, the percent of agreement between any other two models is above $70\%$. Especially, each two of Claude-3.5-sonnet, DeepSeek-R1, and Gemini-1.5-flash have a percent of agreement above $75\%$. Llama-3.1-8b has relatively low agreements with other models.
In the 2-class setting (CC2), the results are similar: high agreement among Claude-3.5-sonnet, DeepSeek-R1, and Gemini-1.5-flash, and low agreement between Llama-3.1-8b and other models. 
A notable difference is that GPT-4o-mini has less agreement with other models in the 2-class setting than in the 4-class setting. 
Also, DeepSeek-V3 agrees with DeepSeek-R1 with most of the entries in classification (82\% in the 4-class setting and 84\% in the 2-class setting).

\spacingset{1.2}
\spacingset{1.45}

 In summary, for the four settings we have considered so far (AC1, AC2, CC1, and CC2),  DeepSeek underperforms Claude, but consistently outperforms Gemini, GPT, and Llama.  
Also, DeepSeek is computationally slower than others, and Claude is much more expensive than others. 
Given that DeepSeek is a new LLM where the training cost is only 
a fraction of that of other LLMs, we expect that in the near future,  DeepSeek will grow substantially and may  become   the most appealing LLM approach for our study.

\subsection{Temporal stability of LLMs} \label{subsec:stability}
Since an LLM is as a black box to users and may have randomness in its output, it is crucial to assess the stability of its performance across multiple runs at different time points.
We investigate this by using the citation classification problem for example. 

First, we examine the short-term temporal variation of 6 LLMs.
We randomly selected 5\% of samples (149 citation instances) and repeated Experiments CC1 and CC2 for 5 times, with a 6-hour separation between two nested runs.   
The min, mean, and maximum error rates among 5 runs are in Table~\ref{tb:stability-short-run}. Except for Llama-3.1-8b, the other LLMs have stable error rates across multiple runs, where the maximum difference is no more than $0.04$ in the 4-class setting and $0.05$ in the 2-class setting. These variations have no influence on the performance rank among these 4 LLMs. Llama-3.1-8b has a large variation, especially in the 2-class setting, in which the maximum error rate is nearly twice of the minimum one.

\spacingset{1.2}
\begin{table}[h!]
\centering
\scalebox{1}{
\begin{tabular}{l | ccc  | ccc}
\toprule
\multirow{2}{*}{\textbf{Method}} & \multicolumn{3}{c|}{4 classes} & \multicolumn{3}{c}{2 classes}\\
\cline{2-7}
& \textbf{Min} & \textbf{Mean} & \textbf{Max} & \textbf{Min}& \textbf{Mean} & \textbf{Max}\\
\hline
Claude-3.5-sonnet & 0.322 & 0.326 & 0.336  & 0.221 & 0.250 & 0.268 \\
DeepSeek-V3 & 0.409  &0.420   & 0.430  & 0.315 & 0.325  & 0.342 \\
 DeepSeek-R1  &  0.362    & 0.384 & 0.396 &   0.262   &  0.275  & 0.289 \\
Gemini-1.5-flash  & 0.329 & 0.332  & 0.336  & 0.268 & 0.287  & 0.295 \\
GPT-4o-mini & 0.315 & 0.325  & 0.336 & 0.430 & 0.446  & 0.463\\
Llama-3.1-8b & 0.550 & 0.585  & 0.631 & 0.221 & 0.370  & 0.436\\
\bottomrule
\end{tabular}}
\caption{The short-term variation of 6 LLMs in citation classification. For each LLM, the minimum, mean and maximum error rates are calculated based on 5 separate runs, with a 6-hour interval between consecutive nested runs.}
\label{tb:stability-short-run}
\end{table}
\spacingset{1.45}

Second, we examine the long-term temporal variation. Except those for DeepSeek-R1, the other experiments reported in Table~\ref{tb:CC-results} were performed on February 3, 2025. We re-ran these experiments on February 10, 2025, with one week's difference between the two runs.   The experiments on DeepSeek-R1 were performed on May 8, 2025, and a second run was conducted on May 15, 2025, again maintaining one week's difference between the two runs.  
In Table~\ref{tb:stability-long-run}, the columns under ``Error 1" are copied from Table~\ref{tb:CC-results}, and the columns under  ``Error 2" come from the second run. Except GPT-4o-mini and Llama-3.1-8b, the other LLMs have similar error rates in these two runs. 

We also calculated the percent of agreement on predicted labels between these two runs (when there is high agreement, the two runs must have similar error rates; but the opposite is not necessarily true).  As we can see in Table~\ref{tb:stability-long-run},   DeepSeek-V3 and Gemini-1.5-flash exhibit the highest level of agreement, indicating  strong long-term stability. Claude-3.5-sonnet also has more than 90\% of agreement.  In addition, the agreement of DeepSeek-R1 is approximately 90\%.   For GPT-4o-mini, the percent is between 80\% and 90\%. 
For Llama-3.1-8b, only less than 50\% of predicted labels are the same in two runs.  

\spacingset{1.2}
\begin{table}[h!]
\centering
\scalebox{.9}{
\begin{tabular}{l | ccc  | ccc}
\toprule
\multirow{2}{*}{\textbf{Method}} & \multicolumn{3}{c|}{4 classes} & \multicolumn{3}{c}{2 classes}\\
\cline{2-7}
& \textbf{Error 1} & \textbf{Error 2} & \textbf{Agreement} & \textbf{Error 1}& \textbf{Error 2} & \textbf{Agreement}\\
\hline
Claude-3.5-sonnet & 0.327 & 0.326 &  90.0\% & 0.261 & 0.263 & 91.0\% \\
DeepSeek-V3 & 0.432  & 0.433 & 96.1\%  & 0.332 & 0.331 & 93.5\% \\
 DeepSeek-R1  & 0.380 & 0.384  &  89.1\% & 0.291  &  0.299  &  90.3\% \\
Gemini-1.5-flash  & 0.347 & 0.348 & 97.4\% & 0.313 & 0.314 & 95.3\% \\
GPT-4o-mini & 0.363 & 0.326 & 88.4\% & 0.371 & 0.449 &  80.0\%  \\
Llama-3.1-8b & 0.576  & 0.649 & 46.0\% & 0.457 &  0.489 & 54.2\% \\
\bottomrule
\end{tabular}}
\caption{The long-term variation of 5 LLMs in citation classification. We ran experiments twice, on February 3, 2025 and February 10, 2025,  and reported the respective error rates. ``Agreement" refers to the percent of agreement on predicted labels between two runs.}
\label{tb:stability-long-run}
\end{table}
\spacingset{1.45}

In summary, Claude-3.5-sonnet, DeepSeek-V3, DeepSeek-R1, and Gemini-1.5-flash are stable in terms of both short-term and long-term variations, GPT-4o-mini is moderately stable, and Llama-3.1-8b is relatively unstable.

{\bf Remark 5}:  When we talk about the temporal stability of an LLM, 
we must ensure that the LLM we use at different time points comes from the same 
version: this is important as an LLM may update relatively frequently.  For example, 
we noticed that %the most recent version of GPT-4o-mini 
GPT-4.1 (released in April, 2025) is significantly better than 
GPT-4o-mini, the version we used in February, 2025 (the time we conducted our initial experiments in this paper); additionally, Claude-3.7-sonnet (a recent released version of Claude) 
 significantly improves over Claude-3.5-sonnet (the version we used in February, 2025).

\vspace{-.3cm}

\subsection{Comparison and combination with statistical methods} \label{subsec:HC}
Long before the advent of LLMs, classification had been extensively studied in statistics and machine learning, leading to the development of numerous algorithms. How does the LLM approach compare to these traditional methods? Does it offer any advantages? Moreover, can statistical methods be effectively combined with LLMs? In this subsection, we explore these questions using the authorship classification as an example. 

For human-vs-AI authorship classification, we may represent a text abstract by a feature vector $X\in\mathbb{R}^p$, with $p$ being the vocabulary size and the entries of $X$ recording how many times each word appears in this abstract. 
For this problem, despite that the vocabulary can be large, {\it useful features are Rare and Weak}. 
This becomes a traditional classification setting, where $X$ is the feature vector.

Modern language models are carefully trained to mimic the human writing style. Consequently, AI-generated content has similar word frequencies as human-generated one for the majority of words. There are only a small fraction of words that show differences in its use between human-generated and AI-generated content (e.g., the vocabulary may have thousands of words, but there are less than 100 words being useful features). This is {\it signal rareness}. The differentiating power of any individual useful word is usually very weak, at a magnitude much smaller than the natural variation of word usage in natural languages. Due to this {\it signal weakness}, it is unreliable to determine if an individual feature is useful.

\spacingset{1.1}
\begin{algorithm}[t!]
\caption{Higher Criticism for classification.} \label{alg:HC}
\medskip
\textbf{Training}: Input are word-count vectors $X_1,\ldots,X_{n_1}\in\mathbb{R}^p$ for $n_1$ human-written abstracts, and word-count vectors $Y_1, \ldots, Y_{n_2}\in\mathbb{R}^p$ for $n_2$ AI-written abstracts.   
\begin{itemize} 
\item For $1\leq j\leq p$, let $\pi_j$ be the p-value of a standard two-sample $t$-test to $X_{1j}, \ldots, X_{n_1j}$ and $Y_{1j}, Y_{2j}, \ldots, Y_{n_1j}$. Sort p-values and use $\pi_{(j)}$ to denote the $j$th smallest p-value.   
\item (Tuning-free feature selection) Let  $j^* = \mathrm{argmax}_{1\leq j\leq p} \bigl\{ \frac{j/p - \pi_{(j)}}{\sqrt{(j/p)(1-j/p)}}\bigr\}$. Let $\widehat{S}$ be the collection of $j$ such that the p-value associated with $j$ is no larger than $\pi_{(j^*)}$ (since $\pi_{(j^*)}$ is the $j^*$th smallest p-value, $\widehat{S}$ contains exactly $j^*$ features).  
We divide $\widehat{S} = \widehat{S}_1 \cup \widehat{S}_2$, where $\widehat{S}_1$ (or $\widehat{S}_2$) contains those features whose $t$-statistics are positive (or negative).  
\end{itemize}

\textbf{Testing}: Given the word-count vector $X^*$ of a test abstract, let $z  = X -( \frac{n_1}{n_1+n_2}\bar{X} + \frac{n_2}{n_1+n_2}\bar{Y})$, where $\bar{X}$ and $\bar{Y}$ are the mean vectors of training samples. Classify this abstract as human-written if and only if $\sum_{j\in \widehat{S}_1}z_j-\sum_{j\in\widehat{S}_2}z_j>0$.
\end{algorithm}
\spacingset{1.45}

Higher Criticism (HC) is a popular approach for dealing with Rare/Weak signals \citep{DJ15}. The original HC \citep{DJ04} was proposed for multiple testing. It was later extended to classification \citep{DJ08} and clustering \citep{IFPCA}. Recently, HC has been used in inference for LLMs \citep{xie2024debiasing}. 
The uniqueness of HC compared to other classification algorithms is that it provides a {\it tuning-free feature selection} step. Most feature selection approaches require to choose a threshold. When signals are individually strong, we may use cross-validation or FDR control to find this threshold; but when signals are Rare/Weak, such approaches are unsatisfactory (e.g., see \citep{DJ15}). HC determines a data-driven threshold that is appropriate to use under Rare/Weak signals.  After the feature selection, the HC classifier applies Fisher's linear discriminant analysis on the selected features. 
Details are in Algorithm~\ref{alg:HC}.

{\bf Remark 6} {\it (The vocabulary used by HC)}: It is common to restrict the vocabulary when using word count features. For example, in topic modeling, it is standard practice to remove stop words and low-frequency words \citep{blei2003latent,TextReview}. In contrast, when applying HC for classification in our setting, we adopt a naive dictionary: the full set of all words that appear in the training documents, without any filtering. Although this choice often results in a large $p$, HC is able to automatically select a small subset of informative words to construct the classifier. We adopt this naive approach to avoid manual effort and to ensure that the HC classifier remains entirely free of user intervention.

\spacingset{1.1}
\begin{figure}[tb!]
\centering
\scalebox{.9}{
\fbox{
\begin{minipage}{1\textwidth}
{\it You are a classifier that determines whether text is human-written or AI-edited. You will be given academic abstracts that are either written by human or edited by AI. 

\medskip

Word distribution analysis: 

\begin{center}
\begin{tabular}{cc}
Word & More Common In\\
\hline
\text{[word]} & ChatGPT\\
\text{[word]} & ChatGPT\\
\vdots & \vdots\\
\text{[word]} & Human\\
\end{tabular}
\end{center}

Respond with exactly one word: either `human' for human-written text or `ChatGPT' for AI-written text. Be as accurate as possible. }
\end{minipage}}
}
\caption{The prompt in the hybrid classifier, where [word] represents a word in $\widehat{S}=\widehat{S}_1\cup\widehat{S}_2$ (these word sets are from the HC classifier), and `ChaptGPT' or `Human' indicates whether this word is in $\widehat{S}_1$ or $\widehat{S}_2$.}\label{fig:prompt-hybrid}
\end{figure}
\spacingset{1.45}

In addition to HC, we introduce a hybrid classifier that combines HC with an LLM. It first runs HC to obtain $(\widehat{S}_1, \widehat{S}_2)$ and then codes these word sets into the prompt. Specifically,
we add a two-column table; for each row, the first column is a word in $\widehat{S}=\widehat{S}_1\cup \widehat{S}_2$, and the second column indicates whether this word is more common in human-written content (i.e., in $\widehat{S}_1$) or AI-generated content (i.e., in $\widehat{S}_2$). Figure~\ref{fig:prompt-hybrid} shows this revised prompt. 
We recall that the LLM approach in Section~\ref{subsec:AC} does not use any training documents, but HC does. 
This hybrid classifier offers a way to utilize training documents in the LLM approach. We call this classifier ``A-HC", where A represents the name of the LLM in use.

The way training documents are used in this hybrid classifier is different from how they are used in the in-context learning by an LLM (e.g., see \cite{dong2024survey} for a survey). In-context learning feeds all training documents into the LLM, but our method only provides a selected word list. For the authorship classification problem considered here, we find that our way of using training documents is more efficient.

\begin{figure}[tb!]
    \centering
    \includegraphics[width=.75\textwidth]{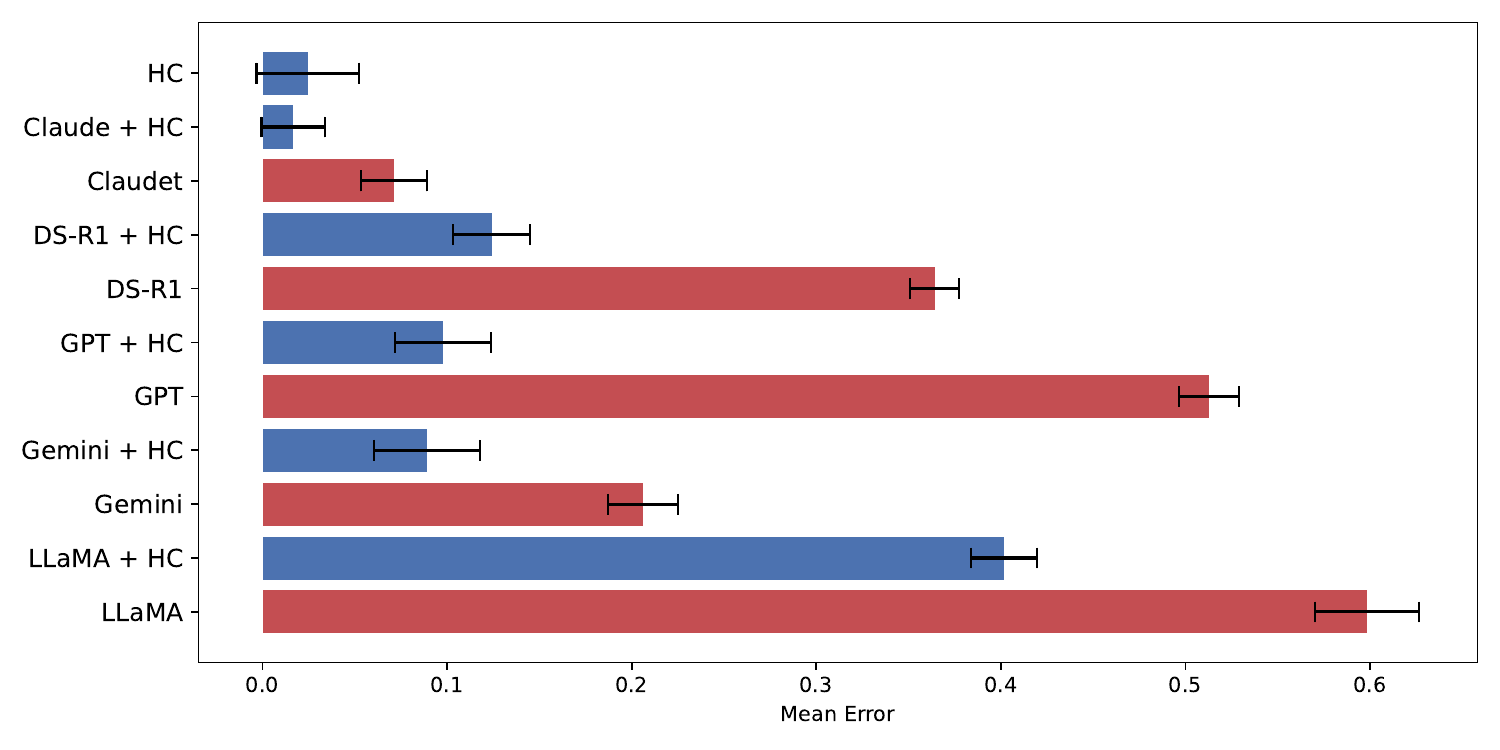}
    \caption{The barplots of per-author classification errors.}
    \label{fig:error-HC}
\end{figure}

We revisited Experiment AC1 (hum vs AI) in Section~\ref{subsec:AC}, using the same dataset of 582 abstract pairs from 15 authors. Since the HC and hybrid classifiers both require training documents, we randomly split each author's abstract pairs into 80\% for training and 20\% for testing. This results in a total of 15 per-author testing error rates for each classifier. The mean and standard deviation are shown in the bar plots in Figure~\ref{fig:error-HC}.   

We have two notable observations: First, the error rates of the HC classifier are significantly lower than those of any pure LLM approaches. One reason is that, even though HC only uses word counts, it has a tuning-free feature selection step to identify the Rare/Weak useful features. In contrast, the pure LLM approach uses numerous hidden  features inside the language model, many of which may be useless for the classification task here.   
Second, if we use the hybrid classifier, then it significantly improves the performance of the original LLM. This improvement is especially large for GPT-4o-mini, where the hybrid version reduces the mean error by 80\%.

Among all the 11 classifiers (HC, 5 LLMs, and 5 hybrid ones), Claude-HC achieves the lowest mean error rate at $0.017$, followed by HC with a mean error rate of $0.025$ and Claude with a mean error rate of $0.071$.  
Among the 5 pure-LLM approaches, Claude is the best, and Gemini is the second. 
DeepSeek-R1 is now ranked the third. The ranks are different from those suggested by Figure~\ref{fig:boxplot}. This is not caused by temporal instability, but the split of training and testing sets --- since each author does not have many abstracts, the error rate on 20\% of randomly selected abstracts could be quite different from the one evaluated on all abstracts.

As the performances of  HC and Claude-HC are similar, we compare the two methods more carefully in a separate experiment. We conducted 5 random splits of each author's abstracts into 80\% for training and 20\% for testing. With a total of $15$ authors, this resulted in $15\times 5=75$ error rates for each method. We computed the mean and standard deviation (SD) of these 75 error rates. For HC, the mean is $5.45\times 10^{-2}$, and the standard deviation is $3.0\times 10^{-2}$. For Claude-HC, the mean is $5.8\times 10^{-3}$, and the standard deviation is $8.7\times 10^{-3}$. These results suggest that Claude-HC significantly improves HC.

The results in this subsection suggest that a traditional statistical method like HC may outperform the pure-LLM approaches. A hybrid of HC and LLM can combine the strengths of both and yield the best performance in the human-vs-AI classification problem.

\subsection{Using other LLMs to generate AI-content} \label{subsec:cross-LLM}

In our previous experiments, the AI-version and humAI-version of each abstract were generated by GPT-4o-mini. In this subsection, we also use Claude-3.7-sonnet and DeepSeek-R1, to generate the AI content. Claude-3.7-sonnet is a new version replacing Claude-3.5-sonnet. This replacement was done after our initial experiments were completed but before this experiment was conducted.  Consequently, the version of Claude in this subsection is different from those in the previous subsections.

We first restricted to the subset of authors who have at least 30 papers in the MADStat data set.  We then randomly selected 15 authors and included all their abstracts (due to randomness, these 15 authors are not the same as in Section~\ref{subsec:AC}). For each abstract, the AI- and humAI-versions were generated using the same prompts as in Section~\ref{subsec:AC}, except that we used three different LLMs, Claude-3.7-sonnet, DeepSeek-R1, and GPT-4o-mini. Fixing one LLM, we pooled together all the included abstracts along with their AI counterparts. 
Finally, we randomly split all abstract pairs into 80\% for training and 20\% for testing. The error rate for each method was calculated by averaging over 5 training/testing splits.

We considered 6 classification methods: three pure LLM approaches and their respective hybrid with HC. 
The results are reported in Table~\ref{tab:cross_llm}. Recent studies show that LLMs often favor their own writing style when judging text \citep{ye2024justice}.  This is not the case when we use the pure LLM approach.
For example, in the top left block of Table~\ref{tab:cross_llm}, the smallest error rate in each column is always achieved by Claude-3.7-sonnet, no matter which LLM is used to generate the AI abstracts; and in the top right block of Table~\ref{tab:cross_llm}, Claude-3.7-sonnet also achieves the best performance in all cases. 
However, when we use the hybrid of an LLM and HC, the results indeed favor having the same LLM for both data generation and classification. 
This is true for both human-vs-AI and human-vs-humAI settings (see the bottom two blocks of Table~\ref{tab:cross_llm}).

\spacingset{1.2}
\begin{table}[htb]
\centering
\scalebox{1}{
\begin{tabular}{l| rrr|rrr}
\toprule
& \multicolumn{3}{c|}{AI} & \multicolumn{3}{c}{humAI}\\ 
\cline{2-7}
& GPT & Claude & DeepSeek & GPT & Claude & DeepSeek \\
\midrule
GPT-4o-mini & 0.500 & 0.486 & 0.500 & 0.500 & 0.438 & 0.500 \\
Claude-3.7-sonnet & 0.372 & 0.204 & 0.388 & 0.438 & 0.312 & 0.438\\
DeepSeek-R1 & 0.418 & 0.352 & 0.390 & 0.459 & 0.322 & 0.500 \\
\hline
HC + GPT-4o-mini  & 0.055 & 0.085 & 0.111 & 0.129 & 0.165 & 0.197 \\
HC + Claude-3.7-sonnet & 0.075 & 0.064 & 0.187 & 0.125 & 0.125 & 0.188 \\
HC + DeepSeek-R1 & 0.132 & 0.203 & 0.058 & 0.312 & 0.375 & 0.125 \\
\bottomrule
\end{tabular}} 
\caption{The error rates of using one LLM for data generation and another for classification. The columns indicate the LLM used to create the AI- or humAI-version of an abstract (to save space, we use abbreviations; e.g., GPT the abbreviation for GPT-4o-mini). The rows indicate the method for classification, where ``HC + LLM" is the hybrid of an LLM with HC (see Section~\ref{subsec:HC} for details).} \label{tab:cross_llm}
\end{table}
\spacingset{1.45}

\vspace{-.5cm}

\section{Discussion} 
 Since the release of its latest version on January 20, 2025,  DeepSeek has been the focus of attention in and beyond the AI community.   It is desired to investigate how it  compares to other popular LLMs. 
In this paper, we compare  DeepSeek with four other popular LLMs (Claude, Gemini, GPT, Llama)  
using the task of predicting an outcome based on a short piece of 
text.  We consider two settings, an authorship 
classification setting and a citation classification setting.  
In these settings, we find that in terms of the prediction accuracy,  DeepSeek outperforms Gemini, GPT, and Llama in most cases, but consistently underperforms Claude. 

Our work can be extended in several directions. First, it is desirable 
to compare these LLMs with many other tasks (e.g., natural 
language processing,  computer vision, etc.).    For example, 
we may use the ImageNet dataset \citep{ImageNet} to compare these LLMs and see which AI is more accurate in classification. 
Second, for both classification problems we considered in this paper,  it is of interest to further improve the performance of the LLM by combining 
tools in statistics and machine learning. 
We have introduced a hybrid classifier between HC and an LLM, in which the selected word set by HC is passed to the LLM via a prompt. Currently, the selected features are all bag-of-words features. We may use HC to select other useful features, such as n-gram features and sentence parsing features, and feed them into the LLM by a carefully written prompt. Last but not the least,  the datasets we generated can be used not only as a platform to compare 
different approaches, but also as useful data to tackle many interesting problems.  
For example,  the MadStatAI  dataset can be used to identify the patterns of AI generated documents, and the CitaStat dataset can be used to tackle problems such as author ranking or estimating the research interest of an author (see for example \cite{JBES} and \cite{TextReview}).  

\medskip

\noindent
{\bf Data availability statement}: The data sets and code files that support the findings of this study are contained in the supplementary material and also publicly available at \url{https://github.com/ZhengTracyKe/DeepSeek-and-other-LLMs}. 
%These data sets were derived from the following resources available in the public domain: \url{https://dataverse.harvard.edu/dataset.xhtml?persistentId=doi:10.7910/DVN/YIXS6B}. 
%After the paper is accepted for publication, the authors plan to make the code and the data sets publicly available. 

\medskip

\noindent
{\bf Acknowledgements}: The authors thank the editor and the anonymous reviewers for their valuable comments. %This research did not receive any specific grant from funding agencies in the public, commercial, or not-for-profit sectors.
%J. Jin is supported in part by the National Science Foundation Grant DMS-2015469. 
Z.T. Ke is supported in part by Sloan Research Grant FG-2023-19970. 

\medskip

\noindent
{\bf Disclosure statement}: The authors report there are no competing interests to declare.

\spacingset{1}
\small 
%%%%%%%%%%%
%%%%%%%%%%%
%%%%%%%%%%%
%%%%%%%%%%%   
\bibliographystyle{chicago}
\bibliography{refs.bib}

\end{document}